# Utilizing deep learning models for the identification of enhancers and super-enhancers based on genomic and epigenomic features

Zahra Ahani, Moein Shahiki Tash, Yoel Ledo Mezquita and Jason Angel

*Instituto Politécnico Nacional (IPN), Centro de Investigación en Computación (CIC), Mexico city, Mexico*

**Abstract**

Super-enhancers are a category of active super-enhancers densely occupied by transcription factors and chromatin regulators, controlling the expression of disease-related genes and cellular identity. Recent studies have demonstrated the formation of complex structures by various factors and super-enhancers, particularly in various cancers. However, our current knowledge of super-enhancers, such as their genomic locations, interaction with factors, functions, and distinction from other super-enhancers regions, remains limited. This research aims to employ deep learning techniques to detect and differentiate between super-enhancers and enhancers based on genomic and epigenomic features and compare the accuracy of the results with other machine learning methods In this study, in addition to evaluating algorithms, we trained a set of genomic and epigenomic features using a deep learning algorithm and the Python-based cross-platform software to detect super-enhancers in DNA sequences. We successfully predicted the presence of super-enhancers in the sequences with higher accuracy and precision.

**Keywords**
Super-enhancer, Enhancer, Genomic, Epigenomic, Deep learning

## 1. Introduction

Artificial intelligence has long been an active field in computer science research. The concept of intelligent computers emerged with the introduction of programmable computers. One of the earliest approaches in artificial intelligence was the use of neural networks and deep learning [1]. In recent years, with increased hardware processing power, machine learning and deep learning methods have rapidly expanded their applications across various scientific domains [2]. The abundance of DNA sequence data and the presence of hidden patterns within this data have led to the convergence of bioinformatics and deep neural networks. These networks effectively handle large datasets and uncover valuable features and hidden relationships within the data.[3]

The identification of super-enhancers in DNA sequences has always been a prominent issue in bioinformatics, with various methods proposed for prediction. Given the common ground

z.ahani2023@cic.ipn.mx (Z. Ahani); mshahikit2022@cic.ipn.mx (M. S. Tash)

between bioinformatics and deep neural networks, this research introduces a new deep learning-based model to assist in finding and predicting super enhancers within DNA sequences. In recent years, deep learning models have seen a resurgence in use [4]. Notably, deep artificial neural networks have achieved significant success in various pattern recognition and machine learning competitions.

Enhancers are regions within DNA strands that not only enhance the transcription of related genes but also play a pivotal role in the expression of specific cell types [5]. Numerous factors are connected to these enhancers, controlling gene expression through the activation of RNA polymerase II and activators [6, 7]. Given their significant role in diseases like diabetes, Alzheimer's, and especially cancer [8], understanding the workings of super-enhancers is of utmost importance.

In the realm of bioinformatics, one of the most prevalent non-laboratory approaches to identifying different regions within aligned DNA sequences is the utilization of algorithms and machine learning techniques. Deep learning, as a cutting-edge and highly effective machine learning algorithm, has found wide-ranging applications across various bioinformatics domains. Examples include sequence analysis [9, 10], structure prediction [11, 12], performance prediction [13], and biomolecule binding prediction [14, 15]. Notably, deep learning outperforms linear regression-based methods in gene expression prediction, particularly when incorporating epigenetic data, which enhances predictive accuracy. The deep code model, which incorporates both genomic sequences and epigenetic features, has demonstrated the superiority of epigenetic attributes in predicting various patterns, including disease. Previous research has consistently shown that gene expression prediction based on epigenetic data is more accurate than genomic data-based predictions. Therefore, the integration of these two domains holds the promise of significantly improving gene expression prediction, aiding in disease prediction, diagnosis, and the development of epigenetic and genetic therapies to combat ailments such as cancer and Alzheimer's.

Deep learning networks, characterized by additional hidden layers, offer an advantage in capturing complex correlations between input features, which is crucial in biological and medical studies. Given the discovery of super-enhancers, researchers have employed sequencing technologies to identify and predict super-enhancers in various cell types. Computational methods have emerged as a priority for super-enhancer prediction due to the challenges and costs associated with laboratory techniques, particularly when dealing with multi-layer structures of factors in gene expression. Specific factors contributing to super-enhancer formation remain largely unknown, underscoring the importance of computational approaches in this context."

## 2. Related work

Enhancers and super-enhancers are regulatory elements within the genome that play critical roles in modulating gene expression and cellular functions, with super-enhancers being particularly associated with the regulation of key genes and cell identity.[16]

## 2.1. Works related to enhancers andsuper-enhancers

Ekaterina V. Kravchuk et al (2023) [17]. In this work super-enhancers (SEs), pivotal cis-regulatory elements in the human genome, were thoroughly investigated. The research systematically reviewed the existing literature on SEs, providing insights into their structural composition and functional significance. Additionally, the study explored the potential applications of SEs, encompassing areas like drug development and clinical utilization. Noteworthy findings included the association of SEs with various pathological conditions and suggestions for enhancing SE identification algorithms, with a preference for robust algorithms like ROSE, imPROSE, and DEEPSEN. A notable research focus emerged in the realm of cancer-associated super-enhancers and prospective therapeutic strategies, offering valuable insights for experts in the field.

Yaqiang Cao et al (2023) [18]. investigate the spatial folding of the eukaryotic genome and its role in genome function. They introduce the Hi-TrAC method, designed for detecting chromatin loops within accessible genomic regions, leading to the identification of active sub-TADs. These sub-TADs contain cell-specific genes and regulatory elements like super-enhancers. The study emphasizes the importance of sub-TAD boundaries and demonstrates their impact on chromatin interactions and gene expression. Hi-TrAC proves to be a cost-effective approach for studying dynamic changes in active sub-TADs, providing valuable insights into genome structures and functions.

Marianna Koutsi et al (2022) [19]. delve into the intricacies of non-coding segments within the human genome, focusing on cis-regulatory modules, including enhancers and super-enhancers (SEs). They shed light on the role of genomic variations in cancer pathogenesis, particularly within these regulatory elements, which can disrupt gene expression profiles. The study highlights the mechanisms of gene expression dysregulation in cancer, addresses cutting-edge technologies for analyzing genomic variations, and explores potential pharmacological approaches to counteract SEs' aberrant function. Additionally, an intratumor meta-analysis reveals a correlation between genomic variations in transcription-factor-driven tumors and the overexpression of certain genes, including cancer-related transcription factors. Various enhancer and SE databases are also introduced as valuable resources for annotating the human genome.

## 2.2. Works related to machine learning

In a study conducted by Moein Shahiki et al (2022) [20]. the focus was placed on language identification within code-mix data, specifically in Kannada-English texts, as part of the CoLI-Kanglish 2022 shared task. The task aimed to discern various languages present in the CoLIKanglish 2022 dataset, which encompassed categories such as Kannada, English, Mixed-Language, Location, Name, and others. Employing classification techniques including K-Nearest Neighbors (KNN) and Support Vector Machine (SVM), the study achieved an F1-score of 0.58, securing a commendable third place among nine competitors. Notably, the study utilized text data from YouTube and categorized words into six predefined groups. The research highlighted the potential of KNN and SVM in code-mix language identification, with KNN showing particularly promising results in achieving the highest weighted average F1-score.

In the research conducted by Mesay Gemeda Yigezu et al(2022)[21]. the focus was directed towards code-mixed language identification (LID) with the objective of determining the language within a given speech segment, word, sentence, or document. The study aimed to identify English, Kannada, and mixed languages using the CoLI-Kenglish dataset, which contained relevant words. The research encompassed various experiments to ascertain the most effective model, ultimately implementing the Bidirectional Long Short Term Memory (BiLSTM) model, which exhibited superior performance with an F1-score of 0.61%. Notably, the precision, recall, and F1-score for identifying Kannada and English languages were reported as 0.66, 0.60, and 0.61, respectively, emphasizing the effectiveness of the BiLSTM model with attention mechanisms in addressing the language identification challenge

## 3. Methodologhy

For this study, we considered 12 feature sets (DNA Sequence, HMS, CRs, Lsd1-NuRD, Cohesion, DNaseI, POL II, P-TEFb, Mediator, Coactivators, TFs, Chromatin) from the Gene Expression Omnibus (GEO) as our data source, following the dataset utilized in [22]. These datasets encompass 5,168 super-enhancers and 5,168 non-super-enhancers, designated as positive and negative classes, respectively. Each case is represented by 45 features, which consist of 20 features related to chromatin, 11 for transcription factors, 11 for motifs, and 3 pertaining to specific sequence information.

In this study, we employ two deep learning approaches for super-enhancer prediction:
Simple deep learning method (neural network with additional hidden layers) and One-dimensional convolutional approach.

DPNN incorporates neural networks with hidden layers, while Conv1D utilizes one-dimensional convolution, commonly employed in image and video processing. The choice of models hinges on their effectiveness in handling genomic and epigenomic data to identify super-enhancers

### 3.1. Simple deep learning method

According to the general approximation theorem, if an adequate number of layers (hidden layers) are designed in deep neural networks, and an acceptable margin of error is defined for these algorithms in the field of artificial intelligence and machine learning, these models can approximate, learn, and represent any type of function. Therefore, a neural network can be defined in the form of an information flow from the input layer to the hidden layers and then to the output layer. For a three-layer neural network, the real function to be learned by the network will be represented as follows: $f(x) = f3(f2(f1(x)))$ In other words, in each of the layers of artificial neural networks with multiple layers, different representation models are learned or, in other words, trained. In all machine learning models, including multi-layer artificial neural networks, real functions (meaning real functions in the field of machine learning, whether linear or nonlinear, learned by these methods and used to make decisions about input data) are rarely linear. Since the representation model plays a vital role in ensuring the optimal performance of machine learning models, using artificial neural networks, the system will be able to learn highly complex and data-driven representation models.

## 3.2. Convolution method

One of the most common deep learning methods in solving machine learning problems, widely used in image and video processing, is CNNs[23]. CNNs extract local spatial correlation in the input by local connections between neurons in adjacent layers. A clear three-level architecture of it is shown in Figure 1.

Figure 1: Convolutional Network Training

## 3.3. Determining the model parameters

The primary deep learning methods employed are a Single Hidden Layer Neural Network (DPNN) and one-dimensional convolution using a conv1d filter. Both methods utilize layers with 16 units, except for the output layer, which has 2 units, reflecting the analysis' focus on classifying inputs into the super-enhancer category. Extensive experimentation was conducted on all model parameters, revealing that, aside from the 'epochs' parameter, other adjustments had minimal impact on prediction outcomes. Consequently, default values in the Keras software were adopted for these parameters. The selection of these parameters—layer depth, neuron count, epochs, and filter sizes—was meticulously tailored to address the complexity of the enhancer prediction task and the necessity to capture intricate DNA sequence patterns. Fine-tuning epochs and learning rates aimed to ensure model convergence without overfitting, while optimizing Conv1D filter sizes targeted essential local spatial correlations within input sequences. Drawing from existing studies and empirical evidence, these parameter configurations were carefully crafted to strike a balance between computational efficiency and model robustness, resulting in architectures positioned for promising performance and generalizability.

Figures 1 and 2 correspond to the values tested to find the optimal number of epochs. For both methods, we selected 20 epochs as this was the first value that provided stable and consistent predictions on both training and testing datasets. To construct a neural network, we first import the Cross library[24] and essential modules from kerass: with importkeras[25] Then, it was necessary to import several modules from Cross: the Sequential module is used for initializing the ANN(Artificial Neural Network), and the Dense module is used for creating the layers of the ANN the imports are from keras. models import Sequential and from keras. layers import Dense then, we initialized the ANN by creating an instance of the Sequential class. The Sequential function initializes a linear stack of layers. Layers can be added further on to this stack using the Dense module. classifier = Sequential()

## 3.4. k-fold cross-validation

In employing k-fold cross-validation within deep learning architectures, the traditional separation of data into distinct training, validation, and test sets might be bypassed due to several reasons. Primarily, this approach maximizes data utilization, especially in scenarios with limited datasets, allowing for more robust model evaluation without compromising training size. By repeatedly training on different subsets and averaging performance metrics across folds, k-fold CV inherently integrates validation within its iterative training process, obviating the need for a separate validation split. This strategy enables comprehensive hyperparameter tuning, fosters a deeper understanding of model generalization, and presents a more holistic evaluation framework for the model's performance, acknowledging the trade-offs associated with traditional train/validation/test splits.

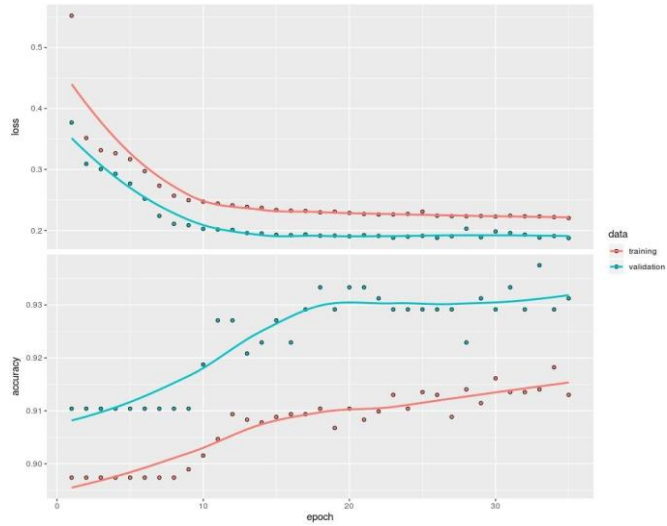

**Figure 1:** Selecting the optimal parameter 'epoch' for the DPNN method

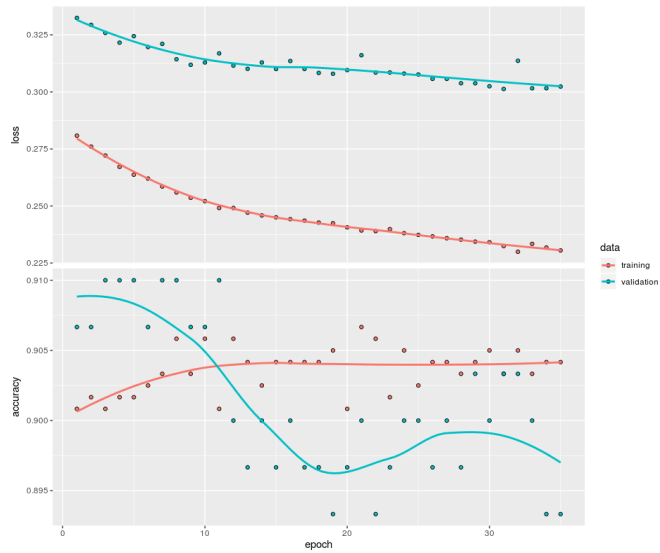

**Figure 2:** Selecting the optimal 'epoch' parameter for the CONV1D method

## 4. Results

### 4.1. Results based on the DPNN method

In the beginning, we considered all features to assess the capability of this method for predicting enhancers using this set of features.K-fold cross-validation with k=10 was employed for this purpose.

Table 1 displays the prediction results for each category based on various metrics.

**Table 1**
Prediction results by the DPNN method using all features

| Folds | Loss | Acc | Precision | Recall | F1-score | Auc |
| --- | --- | --- | --- | --- | --- | --- |
| 1 | 0 | 0.9645 | 0.9064 | 0.806 | 0.9321 | 0.9623 |
| 2 | 0.0801 | 0.9631 | 0.8823 | 0.824 | 0.792 | 0.9601 |
| 3 | 0 | 0.91 | 0.94 | 0.8556 | 0.9606 | 0.9311 |
| 4 | 0.0071 | 0.9577 | 0.8923 | 0.8474 | 0.792 | 0.95 |
| 5 | 0.151 | 0.9606 | 0.8671 | 0.8146 | 0.8465 | 0.973 |
| 6 | 0.2 | 0.9469 | 0.9246 | 0.8224 | 0.8269 | 0.9543 |
| 7 | 0.0246 | 0.9171 | 0.9047 | 0.8408 | 0.8841 | 0.9222 |
| 8 | 0.0948 | 0.98 | 0.8813 | 0.871 | 0.8976 | 0.9801 |
| 9 | 0.0825 | 0.9397 | 0.8844 | 0.8325 | 0.8115 | 0.9644 |
| 10 | 0.0696 | 0.9761 | 0.9332 | 0.8314 | 0.9653 | 0.983 |
| Ave | 0.071 | 0.9516 | 0.9016 | 0.8346 | 0.8709 | 0.958 |

While the results are generally acceptable, for comparison, the feature groups were subjected to analysis under identical conditions using the DPNN method based on genomic and epigenomic classification. The analysis results are presented in Tables 2 and 3.

**Table 2**
Prediction results by the DPNN method using genomic features

| Folds | Loss | Acc | Precision | Recall | F1-score | Auc |
| --- | --- | --- | --- | --- | --- | --- |
| 1 | 0.1638 | 0.91 | 0.94 | 0.835 | 0.83 | 0.932 |
| 2 | 0.0861 | 0.9536 | 0.8163 | 0.7931 | 0.83 | 0.972 |
| 3 | 0.1443 | 0.9797 | 0.9341 | 0.8193 | 0.8556 | 0.9555 |
| 4 | 0.0845 | 0.91 | 0.8915 | 0.79 | 0.83 | 0.9231 |
| 5 | 0.061 | 0.9673 | 0.8762 | 0.8163 | 0.83 | 0.981 |
| 6 | 0.111 | 0.91 | 0.9237 | 0.8466 | 0.84 | 0.9345 |
| 7 | 0.1993 | 0.91 | 0.8932 | 0.7988 | 0.96 | 0.9421 |
| 8 | 0.0685 | 0.9293 | 0.9292 | 0.8555 | 0.9014 | 0.92 |
| 9 | 0.0794 | 0.913 | 0.8407 | 0.8743 | 0.8559 | 0.92 |
| 10 | 0.0257 | 0.9702 | 0.94 | 0.7958 | 0.8636 | 0.9773 |
| Ave | 0.1024 | 0.9353 | 0.8985 | 0.8225 | 0.8596 | 0.9457 |

### 4.2. Comparison of results with the DPNN method across feature groups

To assess the impact of feature groups on prediction, first, the average scores for each evaluation metric were calculated separately for each feature group using cross-validation across all categories. Figure 3 was created to compare the results of this section.

As can be observed, the epigenetic feature group exhibits a significant superiority over the genomic group for all evaluation metrics.

**Table 3**

Prediction results by the DPNN method using epigenomic features

| Folds | Loss | Acc | Precision | Recall | F1-score | Auc |
|---|---|---|---|---|---|---|
| 1 | 0.1817 | 0.9409 | 0.96 | 0.8469 | 0.86 | 0.955 |
| 2 | 0.1149 | 0.9537 | 0.9315 | 0.8741 | 0.86 | 0.9654 |
| 3 | 0.1068 | 0.98 | 0.9211 | 0.8271 | 0.8709 | 0.9765 |
| 4 | 0.0846 | 0.9573 | 0.96 | 0.8415 | 0.942 | 0.9876 |
| 5 | 0.2 | 0.94 | 0.9331 | 0.82 | 0.9151 | 0.9654 |
| 6 | 0.1209 | 0.98 | 0.9184 | 0.8303 | 0.86 | 0.99 |
| 7 | 0.0588 | 0.98 | 0.9179 | 0.8625 | 0.9291 | 0.9766 |
| 8 | 0.1139 | 0.9726 | 0.9299 | 0.8749 | 0.8765 | 0.9874 |
| 9 | 0.1116 | 0.98 | 0.9579 | 0.8515 | 0.8823 | 0.9885 |
| 10 | 0.0511 | 0.9651 | 0.89 | 0.8717 | 0.8668 | 0.9874 |
| Ave | 0.1144 | 0.965 | 0.932 | 0.85 | 0.8863 | 0.9779 |

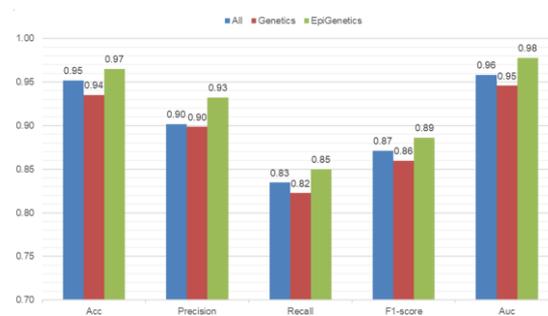

**Figure 3:** Comparison of prediction results using the DPNN method and three feature groups: overall, genomic, and epigenomic

## 4.3. The results based on the Conv1D method

All the analysis steps for the DPNN method were also performed for the Conv1D method. Tables 4, 5, and 6, respectively display the prediction results for the genomic and epigenomic feature groups. Once again k-fold, cross-validation with k=10 and the evaluation metrics used in the previous section were applied.

**Table 4**

Prediction results by the Conv1D method using all features

| Folds | Loss | Acc | Precision | Recall | F1-score | Auc |
|---|---|---|---|---|---|---|
| 1 | 0.1655 | 0.9607 | 0.8897 | 0.8878 | 0.86 | 0.98 |
| 2 | 0.0562 | 0.9323 | 0.8903 | 0.8635 | 0.86 | 0.98 |
| 3 | 0.11 | 0.977 | 0.87 | 0.8181 | 0.86 | 0.98 |
| 4 | 0.12 | 0.93 | 0.91 | 0.89 | 0.88 | 0.97 |
| 5 | 0.0099 | 0.93 | 0.94 | 0.8314 | 0.99 | 0.98 |
| 6 | 0.1724 | 0.98 | 0.9104 | 0.8936 | 0.9227 | 0.98 |
| 7 | 0.1149 | 0.98 | 0.87 | 0.8567 | 0.8615 | 0.98 |
| 8 | 0.1181 | 0.98 | 0.9181 | 0.8465 | 0.86 | 0.98 |
| 9 | 0.0497 | 0.93 | 0.9325 | 0.8486 | 0.86 | 0.98 |
| 10 | 0.0872 | 0.98 | 0.94 | 0.81 | 0.99 | 0.98 |
| Ave | 0.1007 | 0.958 | 0.9076 | 0.8549 | 0.895 | 0.9797 |

**Table 5**
Prediction results by the Conv1D method using genomic features

| Folds | Loss | Acc | Precision | Recall | F1-score | Auc |
|---|---|---|---|---|---|---|
| 1 | 0.0524 | 0.91 | 0.8606 | 0.8428 | 0.8833 | 0.96 |
| 2 | 0.1104 | 0.91 | 0.94 | 0.8557 | 0.83 | 0.9539 |
| 3 | 0.1438 | 0.9696 | 0.881 | 0.8533 | 0.8991 | 0.9507 |
| 4 | 0.0482 | 0.9463 | 0.94 | 0.8 | 0.83 | 0.9476 |
| 5 | 0.1665 | 0.91 | 0.909 | 0.8686 | 0.96 | 0.9471 |
| 6 | 0.1721 | 0.9413 | 0.9085 | 0.8246 | 0.83 | 0.9321 |
| 7 | 0.1286 | 0.974 | 0.8262 | 0.8902 | 0.83 | 0.9289 |
| 8 | 0.1275 | 0.98 | 0.9024 | 0.824 | 0.9283 | 0.9474 |
| 9 | 0.1455 | 0.9241 | 0.8279 | 0.875 | 0.842 | 0.9595 |
| 10 | 0.1314 | 0.9228 | 0.8239 | 0.9187 | 0.9014 | 0.9367 |
| Ave | 0.1226 | 0.9388 | 0.882 | 0.8553 | 0.8734 | 0.9464 |

**Table 6**
Prediction results by the Conv1D method using epigenomic features

| Folds | Loss | Acc | Precision | Recall | F1-score | Auc |
|---|---|---|---|---|---|---|
| 1 | 0.0872 | 0.99 | 0.92 | 0.8955 | 0.9 | 0.99 |
| 2 | 0.0719 | 0.95 | 0.9511 | 0.9164 | 0.99 | 0.986 |
| 3 | 0.0398 | 0.99 | 0.9345 | 0.9307 | 0.9196 | 0.9896 |
| 4 | 0.1081 | 0.99 | 0.99 | 0.87 | 0.9 | 0.99 |
| 5 | 0.1087 | 0.99 | 0.9338 | 0.933 | 0.9 | 0.99 |
| 6 | 0.1042 | 0.95 | 0.92 | 0.87 | 0.9224 | 0.99 |
| 7 | 0.0922 | 0.99 | 0.92 | 0.9178 | 0.9529 | 0.9882 |
| 8 | 0.0691 | 0.99 | 0.99 | 0.9164 | 0.982 | 0.9882 |
| 9 | 0.0661 | 0.9561 | 0.99 | 0.889 | 0.9691 | 0.99 |
| 10 | 0.104 | 0.9818 | 0.9268 | 0.9329 | 0.9648 | 0.99 |
| Ave | 0.0851 | 0.9778 | 0.9476 | 0.9072 | 0.9401 | 0.9886 |

## 4.4. Comparison of results with the Conv1D method across feature groups.

Figure 4 has been created based on the average evaluation metrics across all cross-validation categories using the Conv1D method. It compares the prediction results among the CONV1D method and the three feature groups: overall, genomic, and epigenomic.

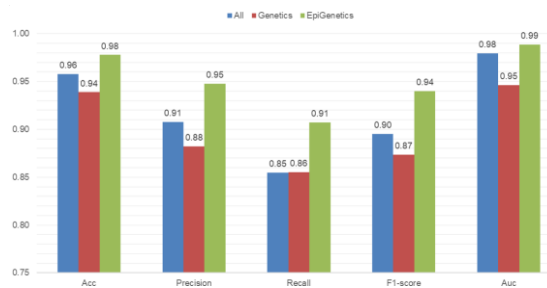

**Figure 4:** Comparison of prediction results based on the CONV1D method and three feature groups: overall, genomic, and epigenomic

The superiority of the epigenomic feature group over the other groups for predicting enhancers is evident once again.

## 4.5. Comparison of the DPNN method with the Conv1D method

In this section, the performance of the two methods, DPNN and Conv1D, for predicting enhancers is examined. For this purpose, the average values of evaluation metrics across all cross-validation categories were analyzed once again for all three feature groups: overall, genomic, and epigenomic. The results of this comparison are summarized in Figures 5, 6, and 7, respectively, for the genomic and epigenomic feature groups.

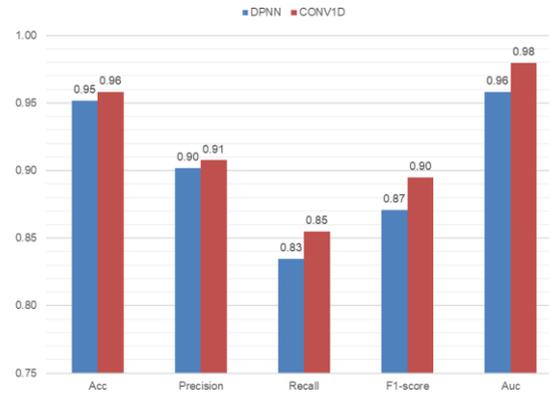

**Figure 5:** Comparison of prediction results between two methods, DPNN and CONV1D, using the overall feature group

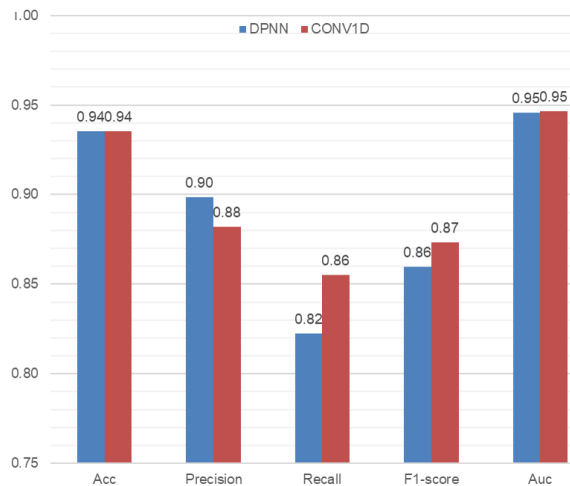

**Figure 6:** Comparison of prediction results between two methods, DPNN and CONV1D, utilizing the genomic feature group

In 2019, Khan and Zhang[22] conducted an analysis using various machine learning methods, including random forest, svm, knn, adaBoost, bayes, treenaive, and decision tree, to predict enhancers on similar datasets. The authors experimented with different parameter settings on various feature groups. The optimal result they achieved was using the random forest algorithm, which had the following evaluation metric scores:

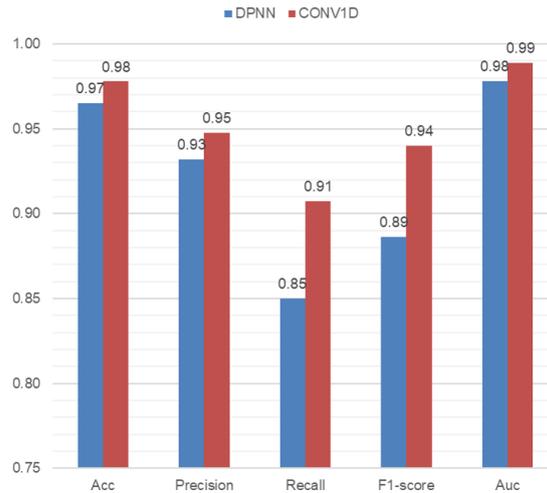

**Figure 7:** Comparison of prediction results for two methods, DPNN and CONV1D, using the epigenomic feature group

- Precision: 0.89 - Recall: 0.82 - F1 Score: 0.85 - AUC (Area Under the Curve): 0.97 While the AUC score for the RANDOM FOREST method may be better in some cases compared to other methods, especially when compared to the DPNN method, overall, their methods had better scores in comparison to other evaluation metrics. This was particularly notable when comparing their results with the best outcome from different analyses, namely, the Conv1D method under the epigenomic feature group.

## 5. limitation

Only two models were used for the data analysis, but there exists the potential to explore additional models for a more comprehensive evaluation. Due to limitations in available datasets, the experiments were restricted to just one dataset. Consequently, there was insufficient space for further comparisons and validations.

## 6. conclusion

In conclusion, this research aimed to leverage deep learning techniques for the detection and differentiation of super enhancers in DNA sequences based on genomic and epigenomic features. The study compared the performance of two deep learning methods, DPNN (Deep Neural Network) and Conv1D (One-dimensional Convolutional Network), across various feature groups.

The results demonstrated that both DPNN and Conv1D methods were effective in predicting super enhancers, with overall good performance across different evaluation metrics. However, when analyzing the feature groups separately, the epigenomic feature group consistently outperformed the genomic feature group, indicating the importance of considering epigenetic information in super enhancer prediction.

Furthermore, the comparison between the DPNN and Conv1D methods revealed that Conv1D performed slightly better on average across all feature groups. This suggests that the one-dimensional convolutional approach may be more suitable for the task of super-enhancer prediction based on the dataset and features used in this study.

Overall, the research highlights the potential of deep learning techniques in the field of bioinformatics and the importance of incorporating epigenomic data for accurate super-enhancer prediction. Further exploration and refinement of these methods could lead to valuable insights into the regulation of gene expression and its implications in various diseases.